\documentclass[runningheads]{llncs}
\usepackage[T1]{fontenc}
\usepackage{bm}
\usepackage{amsmath}
\usepackage{graphicx}
\usepackage{nomencl}
\usepackage{hyperref}
\usepackage{amsfonts}
\usepackage{booktabs} 
\usepackage{siunitx}
\makenomenclature
\newcommand\given[1][]{\:#1\vert\:}

\DeclareMathOperator*{\argmax}{arg\,max}

\begin{document}
\title{POMDP-Based Trajectory Planning for On-Ramp Highway Merging}
%
%
\author{Adam Kollarčík\inst{1,2}\orcidID{0000-0002-9150-839X}\and \\
Zdeněk Hanzálek\inst{2}\orcidID{0000-0002-8135-1296}}
\authorrunning{A. Kollarčík and Z. Hanzálek}
%
\institute{Dept. of Control Engineering, Faculty of Electrical Engineering, Czech Technical University in Prague, Czech Republic \and
Czech Institute of Informatics, Robotics and Cybernetics, Czech Technical University in Prague, Czech Republic\\ \email{\{adam.kollarcik, zdenek.hanzalek\}@cvut.cz}}
\maketitle              
\begin{abstract}
This paper addresses the trajectory planning problem for automated vehicle on-ramp highway merging. To tackle this challenge, we extend our previous work on trajectory planning at unsignalized intersections using Partially Observable Markov Decision \mbox{Processes~(POMDPs)}. The method utilizes the Adaptive Belief Tree (ABT) algorithm, an approximate sampling-based approach to solve POMDPs efficiently. We outline the POMDP formulation process, beginning with discretizing the highway topology to reduce problem complexity. Additionally, we describe the dynamics and measurement models used to predict future states and establish the relationship between available noisy measurements and predictions. Building on our previous work, the dynamics model is expanded to account for lateral movements necessary for lane changes during the merging process. We also define the reward function, which serves as the primary mechanism for specifying the desired behavior of the automated vehicle, combining multiple goals such as avoiding collisions or maintaining appropriate velocity. Our simulation results, conducted on three scenarios based on real-life traffic data from German highways, demonstrate the method's ability to generate safe, collision-free, and efficient merging trajectories. This work shows the versatility of this POMDP-based approach in tackling various automated driving problems.
\keywords{On-Ramp Highway Merging \and Trajectory Planning \and POMDP}
\end{abstract}
\section{Introduction}
\label{introduction}
Highway on-ramp merging is a challenging maneuver for automated vehicles. The ego vehicle must smoothly and safely transition from an on-ramp to the main highway while interacting with other vehicles. The complexity of this task is increased by the need to balance multiple objectives, such as avoiding collisions, maintaining appropriate speed, and completing the merge within the length of the merging lane.

This paper addresses the highway on-ramp merging problem using an approach based on Partially Observable Markov Decision Processes~(POMDPs). POMDPs provide a robust framework for planning and decision-making in scenarios involving uncertainty and incomplete information. As a result, this approach is well-suited to handle the unpredictable behavior of other drivers and the inherent noise in sensor data.

Building on our previous work \cite{vehits24} in automated intersection crossings, we extend the application of POMDPs to the highway merging scenario. This involves enhancing the vehicle dynamics model to account for the lateral movements required to perform a lane change. The simulation results, based on real-world traffic data, demonstrate the effectiveness of the proposed method in producing safe and efficient merging maneuvers.
\section{{Related Work}}
\label{sec:related_work}
Planning approaches for automated vehicles can be divided into three main categories: rule-based, reactive, and interactive methods \cite{eskandarianResearchAdvancesChallenges2021a,hubmannAutomatedDrivingUncertain2018a}. Rule-based and reactive methods do not account for the interconnected behavior of drivers, which is crucial for ensuring safety in automated driving \cite{schwartingPlanningDecisionMakingAutonomous2018a}. As a result, these methods are inadequate for handling complex scenarios such as unsignalized intersection crossings or on-ramp highway merging.

Interactive methods, which can be further categorized into centralized or decentralized, offer a more sophisticated approach by considering the interactions between vehicles \cite{eskandarianResearchAdvancesChallenges2021a}. Centralized methods typically rely on communication between vehicles ({\it Vehicle-to-Vehicle}, V2V), between vehicles and infrastructure ({\it Vehicle-to-Infrastructure}, V2I), or a combination of both ({\it Vehicle-to-Everything}, V2X), to develop a unified global strategy or an informed local strategy \cite{eskandarianResearchAdvancesChallenges2021a,tongArtificialIntelligenceVehicletoEverything2019a}.  While these approaches show promise when most vehicles are equipped for such communication, this is unlikely to be the case in the near future \cite{litmanAutonomousVehicleImplementation2023a}. Therefore, decentralized methods are going to be essential in the coming years. These decentralized approaches can be further divided into three main groups: game theory-based, probabilistic, and data-driven methods \cite{schwartingPlanningDecisionMakingAutonomous2018a}.

In recent years, the on-ramp highway merging problem has been studied thoroughly for the so-called {\it connected and autonomous vehicles} (CAVs), capable of real-time communication \cite{CAV_survey}, fitting into the centralized category. Other approaches, such as \cite{Highways-gameTheory}, use game theory techniques to determine the intentions of other drivers and then use model predictive control (MPC) to generate appropriate trajectories. In \cite{WANG}, the authors provide an in-depth analysis of other highway merging methods, and they present a so-called {\it reference model}, a benchmark model used for verifying automated vehicle technologies. Their reference model utilizes a Monte Carlo tree search and, again, MPC to perform the merge. Additionally, various reinforcement learning methods have been proposed to tackle this problem \cite{RF1,RF2}.

This paper extends our previous work~\cite{vehits24}, in which we implemented and evaluated a probabilistic method based on Partially Observable Markov Decision Processes (POMDPs) for collision-free trajectory planning at unsignalized intersections. Originally introduced in~\cite{hubmannAutomatedDrivingUncertain2018a}, this method has been extended to various scenarios, such as roundabout navigation~\cite{beyPOMDPPlanningRoundabouts2021a} and merging in congested traffic~\cite{Hubman_merge}. Additionally, several enhancements have been proposed, including the consideration of occlusions caused by both static and dynamic objects~\cite{hubmannPOMDPManeuverPlanner2019} and improvements in behavior prediction using dynamic Bayesian networks~\cite{schulzLearningInteractionAwareProbabilistic2019a}. However, to our knowledge, this method has not yet been applied to highway on-ramp merging, as explored in this paper.
\section{Methodology}
To formulate the on-ramp highway merging problem as a POMDP, we begin by providing a formal problem definition. Next, we briefly describe POMDPs and the solver utilized in our approach. Following this, we explain the topology discretization, which reduces the complexity of the problem. Then, we provide a detailed explanation of the models employed for vehicle dynamics and observations and the associated reward function, which encourages collision-free and efficient merging maneuvers.
\subsection{Formal Problem Definition}
\label{subsec::formal_def}
We seek to determine a collision-free trajectory for an ego vehicle during a highway on-ramp merging scenario, where the vehicle must accelerate to achieve a safe velocity before changing lanes to enter a highway.  We assume that the topology of the highway is known beforehand and that the position $\bm{p}_i = [x_i,\ y_i]^\top$, velocity $\bm{v}_i = [v_{x,i},\ v_{y,i}]^\top$, unit heading vector $\bm{\theta}_i$, width $W_i$, and length $L_i$ measurements of all $n$ relevant other (non-ego) vehicles $i \in \{1, \dots, n\}$ are available at every sample time $k$. Our additional assumptions are that (i) non-ego vehicles maintain their lanes, and (ii) that the ego vehicle's state information is known perfectly.

The trajectory of the ego vehicle is defined by a sequence of control inputs: the longitudinal acceleration $a_0^k$ of the ego vehicle (denoted with the zero subscript) and the difference $\Delta \theta^k$ between the ego vehicle's heading and the current lane direction. The trajectory is deemed collision-free if, at every time step $k\geq0$, there is no overlap between the bounding rectangles defined by the ego vehicle’s position $\bm{p}_0^k$, heading vector $\bm{\theta}_0^k$, width $W_0$, and length $L_0$, and the corresponding parameters $\bm{p}_i^k$,  $\bm{\theta}_i^k$, $W_i$, $L_i$ of each non-ego vehicle $i$.

\nomenclature{\(\phi\)}{A path}
\nomenclature{\(t\)}{Parametrization parameter of a path}
\nomenclature{\(d\)}{A path lenght}
\nomenclature{\(n\)}{Number of road users (vehicles)}
\nomenclature{\(\bm{p}\)}{Global position}
\nomenclature{\(\bm{v}\)}{Global velocity}
\nomenclature{\(\bm{\theta}\)}{Global unit orientation vector}
\nomenclature{\(k\)}{Control period (time)}
\nomenclature{\(w\)}{Bounding rectangle width}
\nomenclature{\(l\)}{Bounding rectangle width}

\subsection{POMDPs}
\label{subsec:pomdp}
Partially Observable Markov Decision Processes provide a framework for decision-making and planning under uncertainty~\cite{kurniawatiPartiallyObservableMarkov2022}. A POMDP is defined by the tuple \mbox{$\left< S, A, O, T, Z, R, \gamma\right> $}, where $S$ is the set of states, $A$ is a set of actions, $O$ is the set of observations, $T$ is a set of conditional transition probabilities between states, $Z$ is a set of conditional observation probabilities, $R$ is the reward function, and $\gamma \in \left(0,\ 1 \right]$ is the discount factor. 
The probabilities of transition from state $s \in S $ to state $s'\in S$ with action $a\in A $ are given by $T(s'|a,s)$. Similarly, the probability of observation $o\in O$ in state $s \in S $ with action $a\in A $ is $Z(o|a,s)$.

Due to partial observability, where observations are the only source of information, the knowledge of the current state might be represented by a probability distribution over the set of states, known as {\it{belief}} ${b \in B}$, where $B$ is the {\it belief space}. The belief is updated based on the previous belief $b$, observation $o\in O$, and action $a\in A $, as follows:
\begin{equation}
\label{eq:belief_up}
b'(s') \propto Z(o\given[]a,s')\sum_{s \in S} T(s' \given[] a,s) b(s),\
\end{equation}
where $b(s)$ denotes the probability of being in state $s$ under the belief $b$.

The objective of a POMDP planner is to find a policy $\pi: B\rightarrow A$, that maximizes the expected sum of discounted rewards referred to as the {\it value function}~$V$:
\begin{equation}
\label{eq:val_func}
V = \mathbb{E}\left[\sum_{k=0}^{\infty}\gamma^k R(s_k,a_k)\given[\Big] b,\pi \right].
\end{equation}
To obtain such policy, we use the sampling-based solver {\it TAPIR}~\cite{klimenkoTAPIRSoftwareToolkit2014a}, which implements the {\it Adaptive Belief Tree} algorithm (ABT). This sampling approach does not require us to compute transition or observation probabilities explicitly. Instead, we model the discrete-time stochastic system with difference equations, as detailed in Sections \ref{subsec:models} and \ref{subsec:obs}.
\subsection{Adaptive Belief Tree Algorithm}
\label{subsec:ABT}
The Adaptive Belief Tree algorithm \cite{kurniawatiOnlinePOMDPSolver2016} is an online and anytime algorithm that employs Monte Carlo tree search to find approximate solutions for POMDPs. 
A belief tree $\mathcal{T}$ is constructed by sampling an initial belief with $n_\mathrm{par}$ particles, which are then propagated by actions according to the transition (dynamics) and observation (measurement) models.
At each step, the particles are also assigned a reward \mbox{$r= R(s,a)$}, generating sequences of $N$ quadruples $(s,a,o,r)$ representing state trajectories, where $N$ is the depth parameter of the belief tree. Those trajectories, referred to as {\it episodes}, are also assigned a heuristically computed expected reward for all future states until a terminal state is reached.

Once the tree is constructed, the policy is selected to maximize the average reward of all episodes $h \in H(b,a)$ that include the action-belief pair $(b,a)$ at depth~$l$:
\begin{equation}
\label{eq:policy}
\pi(b) = \argmax_{a\in{\mathcal{T}(b)}}{\frac{1}{|H(b,a)|}\sum_{h\in H(b,a)}\left(\sum_{i = l}^N{\gamma^{i-l}r_i}\right)} ,\
\end{equation}
The average reward is also used to select the actions for particle propagation. 
Initially, actions not yet selected for the current belief are chosen uniformly at random. When there are no unused actions left, the {\it upper confidence bound} (UCB) is employed to address the exploration-exploitation trade-off:
\begin{equation}
a_{\mathrm{sel}} = \argmax_{a \in A}{\left[\frac{1}{|H(b,a)|}\sum_{h\in H(b,a)}\left(\sum_{i = l}^N{\gamma^{i-l}r_i}\right) + c\sqrt{\frac{\log{\sum_a|H(b,a)|}}{|H(b,a)|}}\right]},\
\end{equation}
where $c$ is the tuning parameter for the UCB.

After applying the action $\pi(b_0)$ and obtaining new measurements, the belief is updated. Also, the corresponding branches of the belief tree are reused, with the current belief becoming the new root of the tree. This process is repeated until the goal (terminal state) is reached. To avoid particle depletion, new particles are generated if needed to maintain the $n_\mathrm{par}$ particles.

\subsection{Topology Discretization}
\begin{figure*}[!t]
  \centering
   \includegraphics[width=\textwidth]{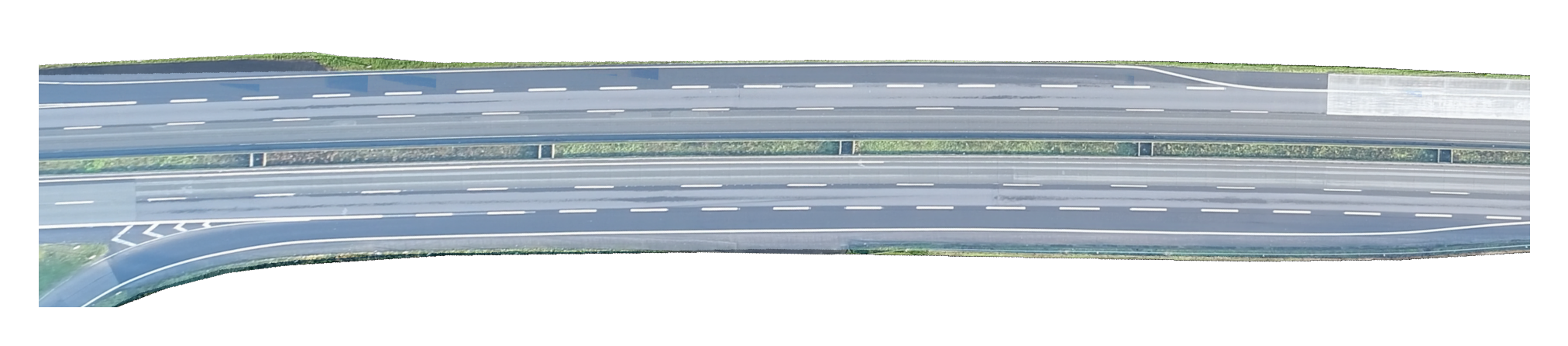} 
    \includegraphics[width=\textwidth]{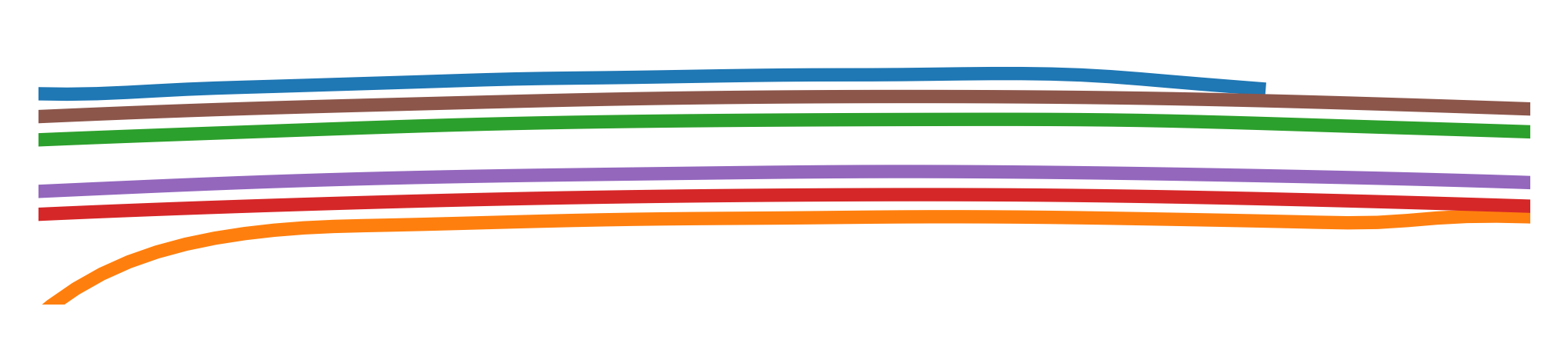}
  \caption{Picture of a highway \cite{exiDdataset} (up) and discrete set of corresponding lanes (down).}
  \label{fig:topo}
 \end{figure*}

\begin{figure}[!t]
\centering
 \includegraphics[width=\textwidth]{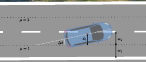}
  \caption{A diagram of a lane change illustrating certain parameters and states. }
  \label{fig:lanesDiagram}
 \end{figure}

Following our original paper's approach, the highway is discretized into a set of $m$ lanes as illustrated in Figure \ref{fig:topo}. Each lane is assigned a unique integer identifier $\mu \in \{1, \dots, m\}$ allowing us to reduce the state description of non-ego vehicles to a triplet of values $\bm{s}_i = [p_{i}\,,\ v_{i} \,,\ \mu_{i}]^\top$, where $\mu_i$ is the lane the $i$-th vehicle is using, $p_i$ is the position along that lane, and $v_i$ is its longitudinal velocity. To allow the ego vehicle to move in the lateral direction, we must extend its state description $\bm{s}_0 = [p_{0}\,,\ d_{0}\,,\ v_{0} \,,\ \mu_{0}]^\top$ by a signed distance $d_0$ from the lane $\mu_0$ (see Figure \ref{fig:lanesDiagram}), where the sign determines left (positive) or right (negative) direction with respect to the orientation of the lane.

\label{subsec:topology}
 
The lanes are extracted using the {\it lanelet2} \cite{poggenhansLanelet2HighdefinitionMap2018} map from the ExiD \cite{exiDdataset} dataset in the OSM XML format. Each lane is stored as two cubic splines with additional information such as length, width, and neighbor lane relations at each lane segment.
 As a result, we obtain mapping from the lane position to the global coordinates $\bm{\phi}_{\mu}(p): p \rightarrow \mathbb{R}^2$, and mapping to the unit heading \mbox{$\bm{\psi}_\mu(p): p \rightarrow \bm{y} \in \mathbb{R}^2: \| \bm{y}\| = 1$}. 
We employ a projection search algorithm from global coordinates to lane position $\epsilon_{\mu}(\bm{p}): \bm{p} \rightarrow \mathbb{R}$ to determine the nearest point on the lane. 

The global position of the ego vehicle can be computed by combining those mappings:
\begin{equation}
 \bm p = \bm{\lambda}_\mu(p,d)=  \bm\phi_\mu(p) +  d \begin{bmatrix}0 & -1 \\ 1 &\phantom{-}0\end{bmatrix} \bm{\psi}_\mu(p),
\end{equation}
where the matrix multiplication of the heading vector generates a perpendicular vector to the path at the position $p$. The signed distance from a lane at a global position $\bm p$ can be computed similarly as well:
\begin{equation}
d = d_\mu(\bm p) = \bm{\psi}_\mu(p)^\top\begin{bmatrix}0 & -1 \\ 1 &\phantom{-}0\end{bmatrix}\bm{p} - \bm{\phi}_\mu(\epsilon_{\mu}(\bm{p})).
\end{equation}
\subsection{Dynamics Model}
\label{subsec:models}
In this section, we present our dynamics model that approximates how the states of the vehicles evolve over time. For the non-ego vehicles, we use a simple single-mass model that effectively describes their movement along the predefined lanes based on their acceleration or deceleration. The mathematical description is as follows:
\begin{align}
\label{eq:modelNonEgoFirst}
p_i^{k+1} &= p_i^k + v_i^k \Delta  t + \frac{1}{2}a_i{\Delta t}^2 +  \nu_{1},      \ \quad i \in \{1, \dots, n\}, \\
v_i^{k+1} &= v_i^k + a_i{\Delta t} + \nu_{2} ,  \qquad  \qquad     \qquad  i \in \{1, \dots, n\}, \\
\label{eq:modelNonEgoLast}
\mu_i^{k+1} &= \mu_i^k, \qquad  \qquad  \qquad  \qquad  \qquad   \ \quad i \in \{1, \dots, n\},
\end{align}
where the $\Delta t$ is the time between subsequent samples $k$ and $k+1$. The Gaussian noise variables $[\nu_1 \,, \nu_2]^\top \sim  \mathcal{N}(\bm{0},\bm{Q})$ with zero mean and covariance $\bm{Q}$ represent inaccuracies in the model. 

For the ego vehicle, we utilize a discretized Frenet frame point model \cite{FRENET}, allowing for the lateral movement necessary to perform a lane change. First, we define auxiliary variables $\hat{p}_{0}$ and $\hat{d}_{0}$ representing the position and signed distance in the next step if no lane change occurred:
\begin{align}
 \hat{p}_{0}&= p_0^k +  \frac{v_0^k \cos{(\Delta \theta^k)} }{1-d_0^k \kappa_{\mu_0}(p_0^k)}\Delta t,\\
\hat{d}_0 &= d_0^k +  v_0^{k}\sin{\left(\Delta\theta^k\right)} \Delta t,
\end{align}
where $\kappa_{\mu_0}(p_0^k)$ is the signed curvature of the lane at position $p_0^k$. Then we check whether the vehicle exceeds the width $w_{\mu_0^k}$ of the lane $\mu_0^k$. If so, and there is a corresponding neighbor lane, a lane change occurs, and state variables are changed accordingly:
\begin{align}
\label{eq:modelEgoFirst}
p_0^{k+1} &=\begin{cases} \epsilon_{\mu_\text{left}}(\bm{\lambda}_{\mu_0^k}(\hat{p}_{0} , \hat{d}_0)) &\text{if} \ \ \phantom{-} \hat{d}_0 > w_{\mu_0^k}( \hat{p}_{0}) \ \text{and} \ {\mu_0^k} \ \text{has left at} \ \hat{p}_{0}, \\ 
 \epsilon_{\mu_\text{right}}(\bm{\lambda}_{\mu_0^k}(\hat{p}_{0} , \hat{d}_0)) & \text{if} \ -\hat{d}_0 > w_{\mu_0^k}( \hat{p}_{0}) \ \text{and} \ {\mu_0^k} \ \text{has right at} \ \hat{p}_{0}, \\   \hat{p}_{0}   & \text{otherwise},\end{cases}\\
 d_0^{k+1} &=\begin{cases} d_{\mu_\text{left}}(\bm{\lambda}_{\mu_0^k}(\hat{p}_{0} , \hat{d}_0)) &\text{if} \ \ \phantom{-}\hat{d}_0 > w_{\mu_0^k}( \hat{p}_{0}) \ \text{and} \ {\mu_0^k} \ \text{has left at} \ \hat{p}_{0},\\ 
  d_{\mu_\text{right}}(\bm{\lambda}_{\mu_0^k}(\hat{p}_{0} , \hat{d}_0)) & \text{if} \ -\hat{d}_0 > w_{\mu_0^k}( \hat{p}_{0}) \ \text{and} \ {\mu_0^k} \ \text{has right at} \ \hat{p}_{0}, \\   \hat{d}_0   & \text{otherwise},\end{cases}\\
  v_0^{k+1} &= v_0^k + a_0{\Delta t},\\
 \mu_0^{k+1} &= \begin{cases} \mu_\text{left} \qquad  \qquad  \qquad \ & \text{if} \  \ \phantom{-} \hat{d}_0 > w_{\mu_0^k}( \hat{p}_{0}) \ \text{and} \ {\mu_0^k} \ \text{has left at} \ \hat{p}_{0},\\ \label{eq:modelEgoLast}
\mu_\text{right}  & \text{if} \ -\hat{d}_0 > w_{\mu_0^k}( \hat{p}_{0}) \ \text{and} \ {\mu_0^k} \ \text{has right at} \ \hat{p}_{0}, \\ \mu_0   & \text{otherwise}.\end{cases}
\end{align}

Furthermore, we need to predict the accelerations for the non-ego vehicles. For the sake of simplicity, we follow our previous approach and employ the so-called intelligent driver model (IDM) \cite{treiberCongestedTrafficStates2000}:
\begin{align}
\label{eq:IDMfirst}
a^k_i = a_\textrm{max} \left[ 1 - \left( \frac{v_i^k}{v_\textrm{des}+\omega_1}\right)^\delta - \left(\frac{d^{*}(v^k_i,v_\textrm{lead})}{d_\textrm{lead}}\right)^2 \right] + \omega_2\,,\quad &i \in \{1, \dots, n\} \\ 
d^{*}(v^k_i,v_\textrm{lead}) = d_\textrm{min} + v^k_i\tau + \frac{v^k_i(v^k_i-v_\textrm{lead})}{2\sqrt{a_\textrm{max} |a_\textrm{min}|}}, \qquad &i \in \{1, \dots, n\}
\end{align}
where $a_\textrm{max}$ is the maximal acceleration, $a_\textrm{min}$ is the minimal acceleration (maximal deceleration), $v_\textrm{des}$ is the desired velocity, $d_\mathrm{min}$ is the minimal distance, $\tau$ is the time headway, $\delta$ is the acceleration exponent,  $v_\textrm{lead}$ is the velocity of the leading (approached) vehicle, and $[\omega_1\,, \omega_2]^\top \sim \mathcal{N}(\bm{0},\bm{\Sigma})$ is Gussian noise with variance $\bm{\Sigma}$ accounting for model inaccuracy and variance in desired velocity values for different drivers. 
The acceleration $a_0$ of the ego vehicle and heading angle deviation $\Delta \theta$ are obtained from the POMDP policy~\eqref {eq:policy}. 
\subsection{Observation Model}

The observation model describes the accuracy of our measurements and how these measurements relate to the states we defined in the previous sections. This is used to generate the particles and to ensure that our current belief of the state corresponds to the obtained measurements.

Let $\bm{z}_i^k$ denote the observation vector of the $i$-th vehicle at time step~$k$. These observations are acquired through spline mappings, as detailed in Section~\ref{subsec:topology}:
\begin{equation}
\label{eq:measurement_model}
\bm{z}_{i}^k = \begin{bmatrix}\bm{\phi}_{\mu_i^k}(p^k_i)\\ {v^k_i} \bm{\psi}_{\mu_i^k}(p^k_i)\\ \bm{\psi}_{\mu_i^k}(p^k_i) \end{bmatrix} + \bm{\zeta},\qquad  \qquad     \qquad  i \in \{1, \dots, n\},
\end{equation}
where $\bm{\zeta} \sim \mathcal{N}(\bm{0}, \bm{R})$ represents Gaussian observation noise with zero mean and covariance matrix $\bm{R}$, encapsulating the measurement inaccuracy together with the error caused by projecting a two-dimensional position on a curve. 

\begin{figure}[!b]
\centering
 \includegraphics[width=0.495\textwidth]{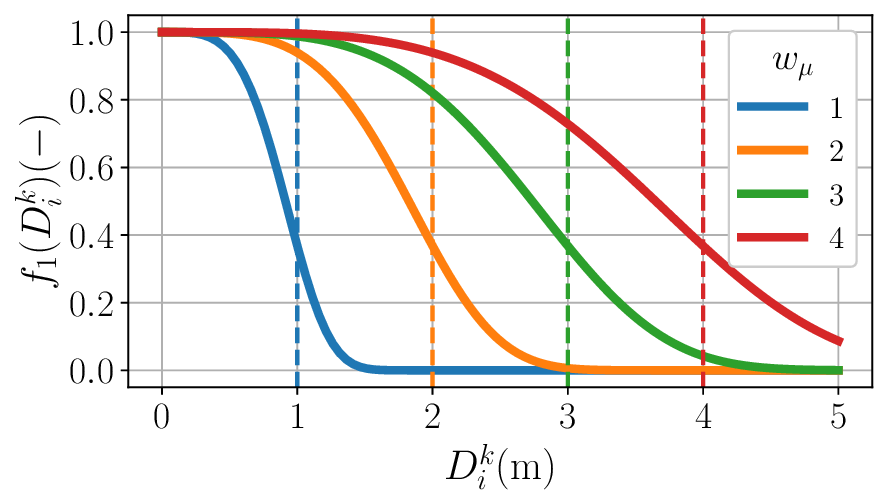}
  \includegraphics[width=0.495\textwidth]{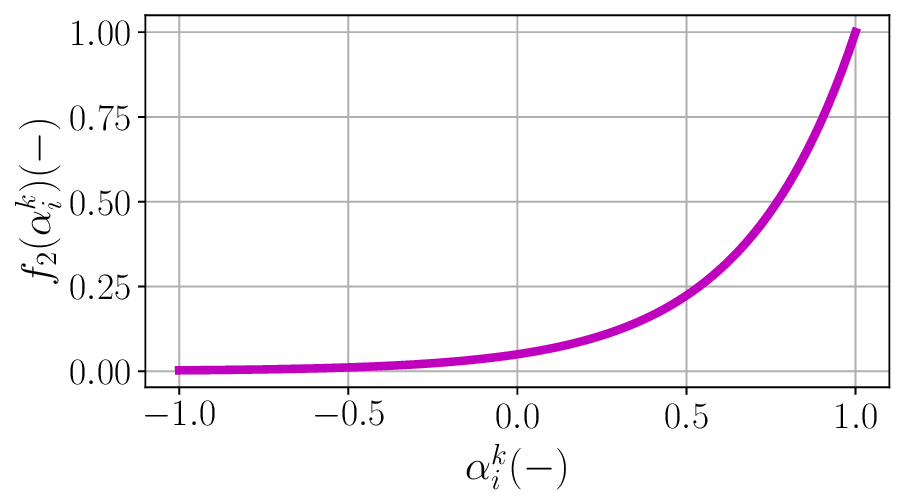}
\caption{Likelihood functions of features $f_1$(left), and $f_2$(right).}
\label{fig:example1}
\end{figure}

Furthermore, we need to estimate the lanes $\mu_i^k$ the vehicles are driving in. In contrast to the planning for unsignalized intersections, highway's topology is simpler having no overlapping segments. This makes the distance from the lane center a good indicator of whether a vehicle is in a particular lane. Nevertheless, additional information, such as vehicle orientation, can also be utilized, as the added complexity is negligible. Thus, we compute the probability of being in a lane for each vehicle based on likelihoods $f_1$ and $f_2$ of these two features shown in Figure~\ref{fig:example1}:
\begin{equation}
    f_1(i,\mu) = e^{-\left(\frac{D^k_i(\mu)}{w_\mu}\right)^4},
\end{equation}
\begin{equation}
    f_2(i,\mu) = e^{3\left(\alpha_i^k(\mu)-1\right)},
\end{equation}
where $D^k_i(\mu) = \|\bm{p}_i^k - \bm{\phi}_{\mu}(\epsilon_{\mu}(\bm{p}_i^k)) \| $ is the distance from the closest point on the lane with width  \mbox{$w_\mu = w_{\mu}(\epsilon_{\mu}(\bm{p}_i^k))$},
 and $\alpha_i^k(\mu) = \bm{\theta}_i^k \cdot \bm{\psi}_{\mu}(\epsilon_{\mu}(\bm{p}_i^k))$ is the dot product of the path heading and the measured heading of a vehicle.
The resulting probability distribution is a combination of these likelihoods:
\begin{equation}
P(i,\mu) = \frac{  f_1(i,\mu)f_2(i,\mu)}{\sum\nolimits_{j=1}^m f_1(i,j)f_2(i,j)}\,.
\end{equation}
\label{subsec:obs}
\subsection{Reward Function and Heuristics}
\label{subsec:reward}
In this section, we describe the reward function and the heuristic used to evaluate the future rewards of newly discovered nodes in the belief tree, as mentioned in Section \ref{subsec:ABT}. The reward function is the primary mechanism for specifying the desired behavior of the ego vehicle, making its formulation critical for achieving our objectives. In conjunction with this, the heuristic is designed to estimate future rewards, guiding the exploration process towards the most promising nodes within the belief tree, thus enhancing the efficiency of the UCB algorithm.

The reward at each time step $r^k$, is composed of five terms, each evaluating a specific requirement. These terms are $ r_\mathrm{vel}$ for velocity, $r_\mathrm{input}$ for control inputs, $r_\mathrm{change}$ for a lane change, $r_\mathrm{crash}$ for collisions, and a direct quadratic component penalizing deviations from the path's centerline with coefficient $R_\mathrm{center}\geq 0$:
\begin{equation}
r^k = r_\mathrm{vel}(\bm{s}_0^k) + r_\mathrm{input}(a_0^k,\Delta \theta^k) +r_\mathrm{crash}(\bm{s}^k) + r_\mathrm{change}(\bm{s}^k) -R_\mathrm{center}\left(d_0^k\right)^2 \, .
\end{equation}

The velocity reward ensures that the ego vehicle follows the desired velocity profile. It is determined by the deviation from the reference velocity \mbox{$ \Delta v = v_\textrm{des} - v^k_0$}:
\begin{equation}
r_\mathrm{vel}(v_0^k) =
\begin{cases}
-R_\mathrm{vel} \Delta v &\textrm{if $\Delta v  \geq 1\,,$}\,\\
-R_\mathrm{vel} {\Delta v}^2 &\textrm{otherwise,}
\end{cases}
\end{equation} 
where $R_\mathrm{vel}$ is a positive weight. 

The input reward function penalizes high acceleration and steering inputs, promoting a smoother ride and increased comfort. It is computed using quadratic penalty function, weighted by $R_\mathrm{acc} \geq 0$ for acceleration and  $R_\mathrm{steer} \geq 0$ for steering:
\begin{equation}
r_\mathrm{acc}(a_0^k,\Delta \theta^k) =
-R_\mathrm{acc} \left({a_0^k}\right)^2 -R_\mathrm{steer} \left({\Delta \theta^k}\right)^2.
\end{equation}

The collision avoidance component is designed to prevent crashes and to keep the vehicle within the highway boundaries. A negative reward $-R_\mathrm{crash}$ is assigned when a collision is detected or if the bounds are exceeded. Otherwise, it is set to zero:
\begin{equation}
r_\mathrm{crash}(\bm{s}^k) =
\begin{cases}
-R_\mathrm{crash}  &\begin{aligned}&\textrm{if collision detected, or}\\ &\textrm{if bounds exceeded,} \end{aligned}\\
0 &\textrm{otherwise.}
\end{cases}
\end{equation} 
Collision detection relies on the overlap of vehicle bounding rectangles, as explained in Section~\ref{subsec::formal_def}. The bounds of the highway are exceeded if
\begin{equation}
\pm \left[d_0 + W_0\cos{(\Delta \theta)} + L_0\sin{(\Delta \theta)}\right] > w_{\mu_0},\,
\end{equation}
while the current lane $\mu_0$ has no left or right neighbor lane.

Finally, the lane change reward $r_\mathrm{change}$  enforces the vehicle to change the line to join the highway. A constant penalty $-R_\mathrm{cst}$ is applied every time step $k$ the vehicle is not in the desired lane. An additional penalty is imposed based on the remaining time $t_\mathrm{left} = (l_{\mu_0}-p_0^k)/v_0^k $ and the remaining distance until the end of the merging lane of length $l_{\mu_0}$. Last, the square distance to the desired lane $d_\mathrm{des} = d_{\mu_\text{des}}(\bm{\lambda}_{\mu_0^k}(p_0^k , d_0^k))$ is also penalized to attract the ego vehicle to the desired lane:
\begin{equation}
r_\mathrm{change}(\bm{s}_0^k) = \begin{cases} -R_\mathrm{cst} - R_\mathrm{end}\left(\frac{1}{t_\mathrm{left}} + \frac{p_0^k}{l_{\mu_0}}\right) -R_\mathrm{dist}(d_\mathrm{des})^2 &\mathrm{if} \, \mu_0^k \neq \mu_\mathrm{des},\\ 0 &\mathrm{otherwise.} \end{cases}
\end{equation}

For the heuristic $h$ estimating the future reward for state $\bm s$, we employed a simple estimation of the minimal time required to reach the desired lane $\mu_\mathrm{desired}$ combined with the crash penalty:
\begin{equation}
h(\bm{s}^k) = -R_h \left|\frac{d_{\mu_\mathrm{desired}}}{v_0^k \sin{\left({\Delta \theta}_\mathrm{max} \right)}}\right| +r_\mathrm{crash}(\bm{s}^k).
\end{equation}
This should guide the UCB algorithm towards states close to the desired lane while avoiding crashes.
\section{{Results}}
\label{sec:results}
Having formulated the problem of on-ramp highway merging as a POMDP, we now present the results, which demonstrate the planning capabilities our method. We ran multiple simulations using three scenarios of three different highway entries from the ExiD dataset \cite{exiDdataset}, which contains over sixteen hours of aerial footage of real-life traffic at highway entries and exits in Germany. In our simulations, we replaced the merging vehicle from the data with our ego vehicle, allowing to reuse the recorded reactions of the other cars as an approximation of interactions with our ego vehicle.

Our simulation software was primarily implemented in C++ using the TAPIR toolbox and ROS1. The simulations were conducted on a laptop with an Intel i5-11500H CPU and showed real-time capable runtimes under our parameter settings. However, it should be noted that the precise real-time performance was not our primary focus and requires further investigation. The parameter values used in our simulations are listed in Table \ref{table:parameters}.
\begin{table}[!h]
\caption{Parameter values for on-ramp highway merging}
\label{table:parameters}
\centering
\begin{tabular}{ccc}
\toprule
 & Description & Value \\ \midrule
$A$            & action set      &     \begin{tabular}{@{}c@{}}$\left\{-1,-0.5,0,0.5,1 \right\}$ \si{\meter \per \square \second}$\times$ \\ $\left\{-2,-1,-0.5,0,0.5,1,2 \right\}$ \si{\deg}\end{tabular}             \\ 
$\Delta t$   & time step length& 1 \si{\sec}\\
$c$            & UCB parameter        & 200000                  \\ 
$N$             & ABT depth       & 10                 \\
$n_\mathrm{ep}$             & ABT number of episodes       & 10000                  \\
$n_\mathrm{par}$              & ABT number of particles       & 1000                  \\
$\gamma$              & discount factor       & 1                 \\
$\bm{Q}$            & dynamics covariance       & $\bm{0} $                 \\
$\bm{\Sigma}$            & IDM covariance       & $\mathrm{diag}(9,0.04)$                 \\
$v_\mathrm{des}$    & desired velocity &  27.8 \si{\meter \per \second} \\
$a_\mathrm{max}$    & IDM maximal acceleration &  1 \si{\meter \per \square \second} \\
$a_\mathrm{min}$    & IDM minimal acceleration &  -1 \si{\meter \per \square \second } \\
$\tau$              & IDM time headway       & 1.5 \si{\second}               \\
$\delta$              & IDM acceleration exponent       & 4              \\
$d_\mathrm{min}$              & IDM minimal distance       & 1 \si{\meter}               \\
$\bm{R}$            & observation covariance       & $\mathrm{diag}(0.01,0.01,0)$                 \\  
$R_\mathrm{center}$            & centerline reward coef.      & 500                  \\
$R_\mathrm{acc}$            & acceleration reward coef.     & 100                  \\ 
$R_\mathrm{steer}$            & steering reward coef.      & $10 \cdot (180/\pi)^2$                  \\ 
$R_\mathrm{vel}$            & velocity reward coef.       & 100                  \\ 
$R_\mathrm{crash}$              & crash reward coef.      & 1000000                 \\ 
$R_\mathrm{cst}$              & wrong lane constant reward coef.        & 1000                \\ 
$R_\mathrm{end}$              & wrong lane end reward  coef.      & 10000                \\ 
$R_\mathrm{dist}$              & wrong lane distance reward coef.        & 500                \\ 
$R_\mathrm{h}$              & heuristics coef.        & 100                \\ 
\bottomrule

\end{tabular}
\end{table}

\subsection{Simulations}
We conducted 30 simulations for each scenario to evaluate the performance of the ego vehicle under varying traffic conditions. Each scenario is illustrated with a time-lapse figure showing the superimposed trajectories of all 30 simulations. We also present velocity and control input graphs to provide further information.

\begin{figure*}[!p]
  \centering
   {\includegraphics[width = \textwidth]{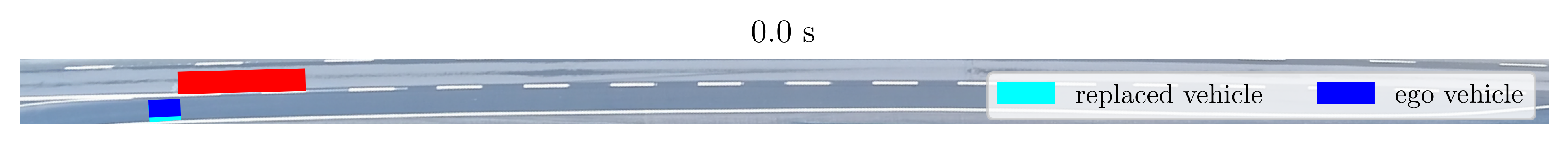}}
    {\includegraphics[width = \textwidth]{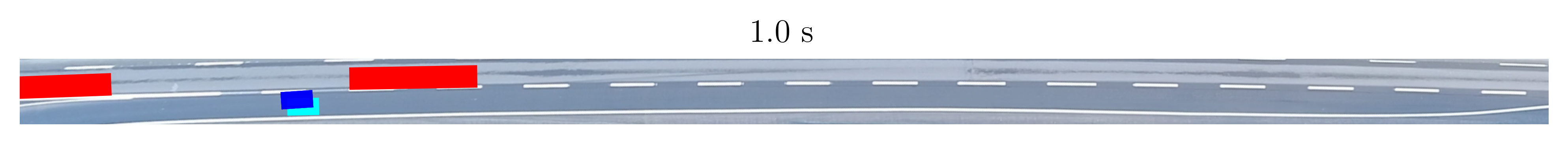}}
    {\includegraphics[width = \textwidth]{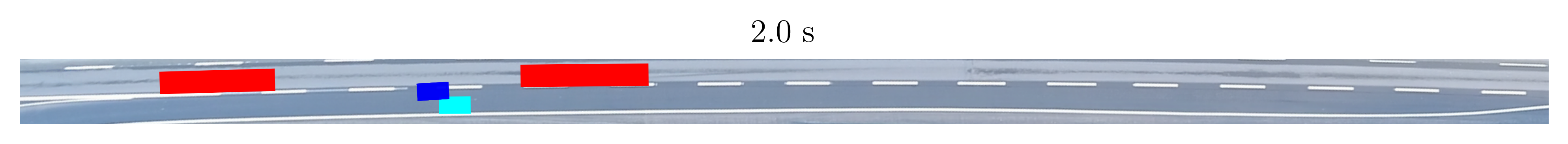}}
    {\includegraphics[width = \textwidth]{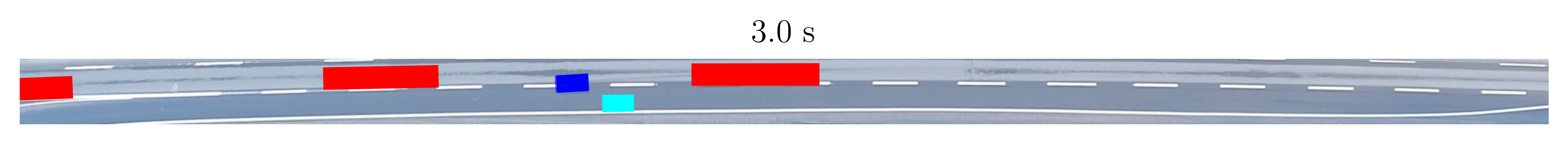}}
    {\includegraphics[width = \textwidth]{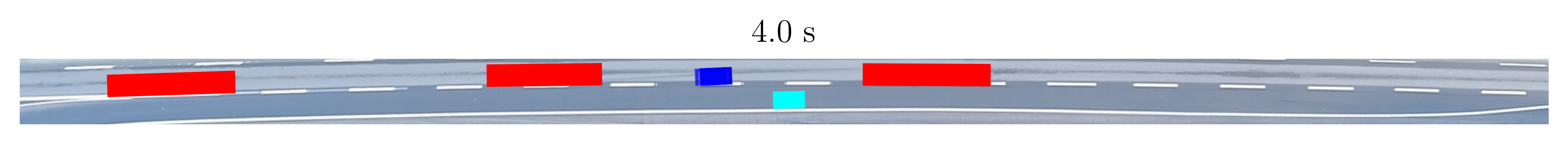}}
    {\includegraphics[width = \textwidth]{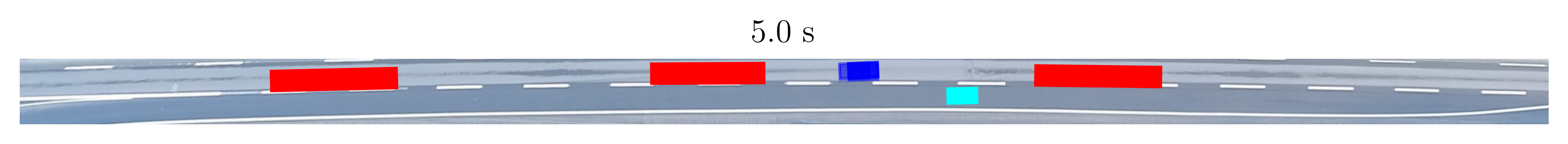}}
    {\includegraphics[width = \textwidth]{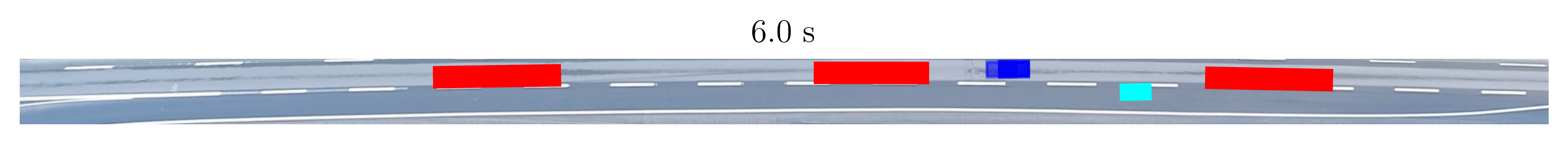}}
    {\includegraphics[width = \textwidth]{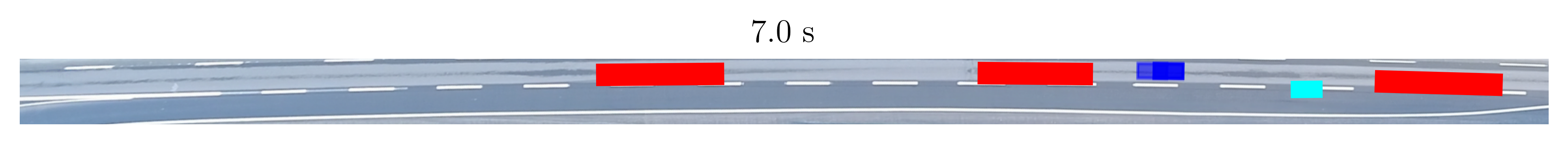}}
     {\includegraphics[width = \textwidth]{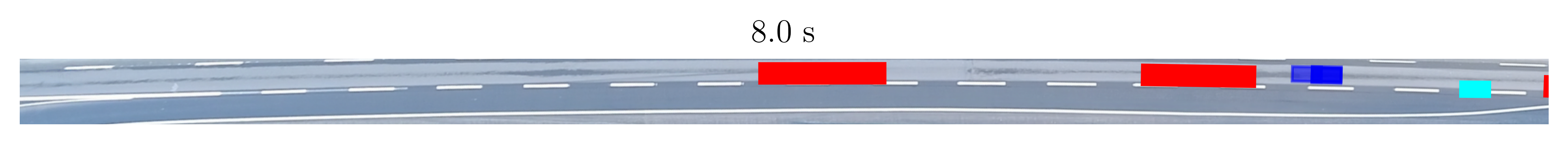}}
    {\includegraphics[width = \textwidth]{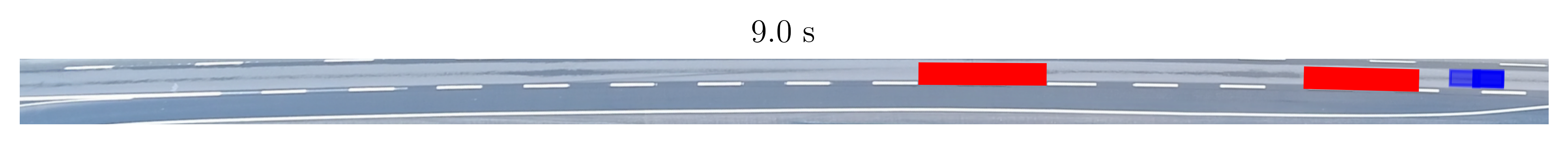}}
  \caption{Time-lapse of the 30 simulations for Scenario 1. Simulated ego vehicles are represented as semi-transparent blue rectangles, non-ego vehicles are shown as red rectangles. The cyan rectangle shows how a real vehicle we replaced with our ego vehicle performed the merge.}
  \label{fig:timelapse1}
 \end{figure*}

\begin{figure*}[!p]
  \centering
   {\includegraphics[width =  \textwidth]{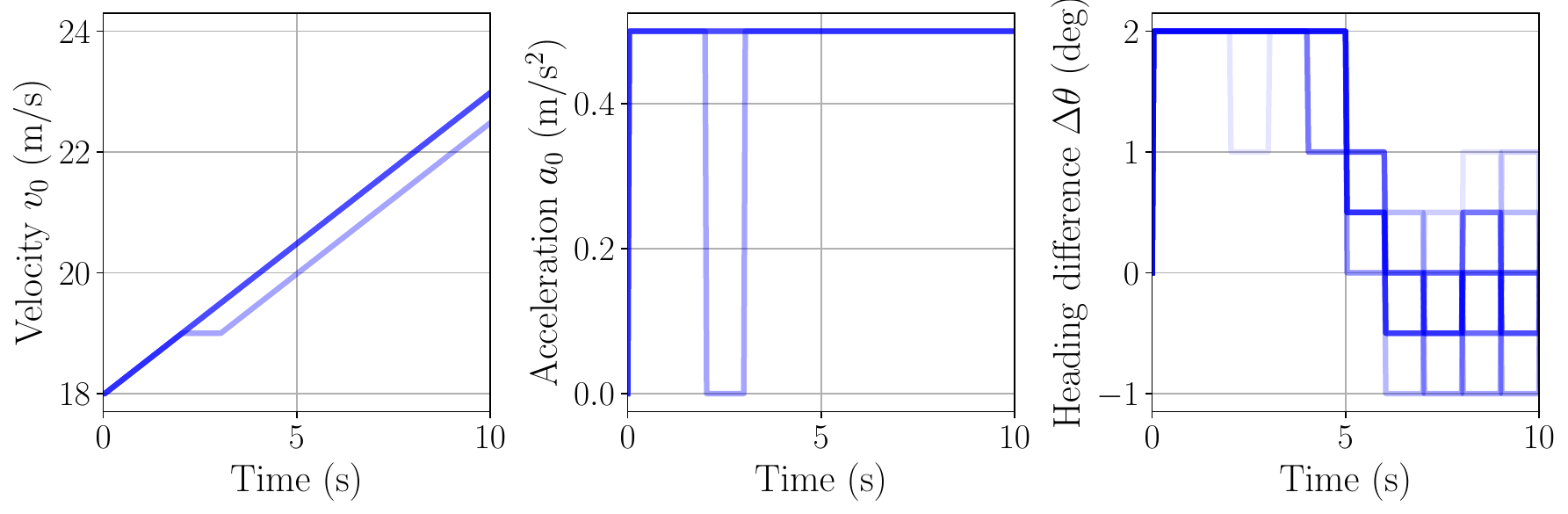}}
  \caption{Velocity and input values of 30 simulations for Scenario 1. The opacity of each line segment represents the total number of simulations sharing the values.}
  \label{fig:values1}
 \end{figure*}

In the first scenario, the ego vehicle initiates a merge onto the highway at a speed of 18 m/s, with three other vehicles present in the target lane. Across all 30 simulations, the ego vehicle successfully changed lanes within 4 seconds, as seen in the time-lapse in Figure \ref{fig:timelapse1}. Although the trailing vehicle appears to get too close post-merge, that is just a consequence of using pre-recorded data. In reality, the trailing vehicle would have had sufficient time to react and slow down; however, because the replaced vehicle in the data was farther away, this adjustment did not occur. As depicted in Figure \ref{fig:values1}, the ego vehicle accelerated uniformly at approximately 0.5 m/$\mathrm{s}^2$ throughout the maneuver.

\begin{figure*}[!p]
  \centering
   {\includegraphics[width = \textwidth]{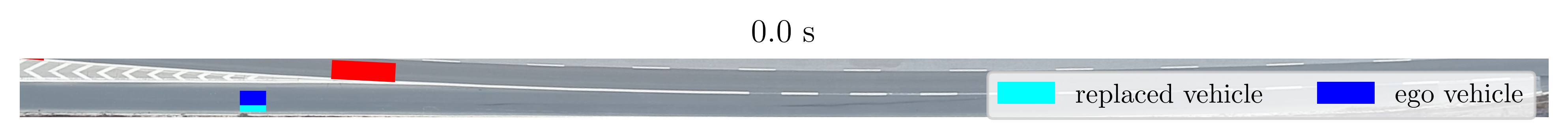}}
    {\includegraphics[width = \textwidth]{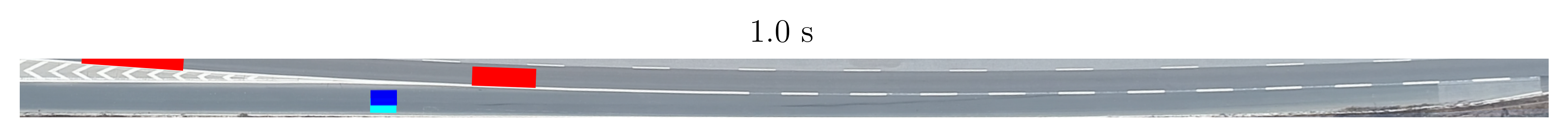}}
    {\includegraphics[width = \textwidth]{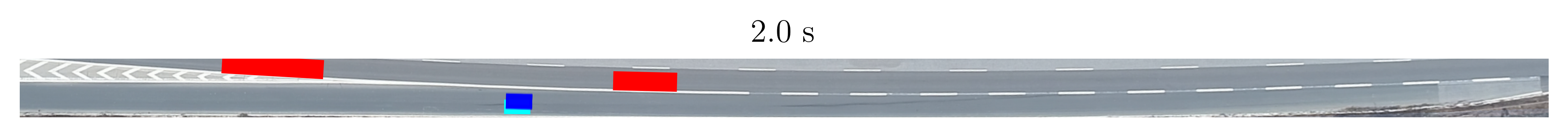}}
    {\includegraphics[width = \textwidth]{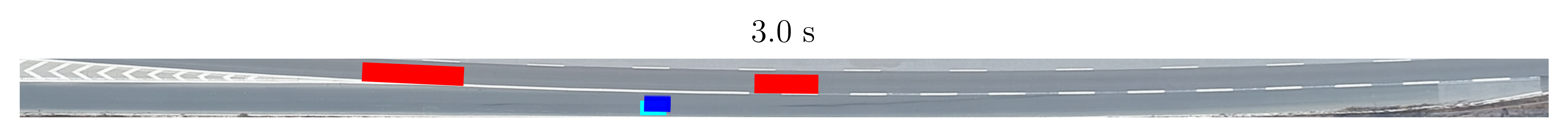}}
    {\includegraphics[width = \textwidth]{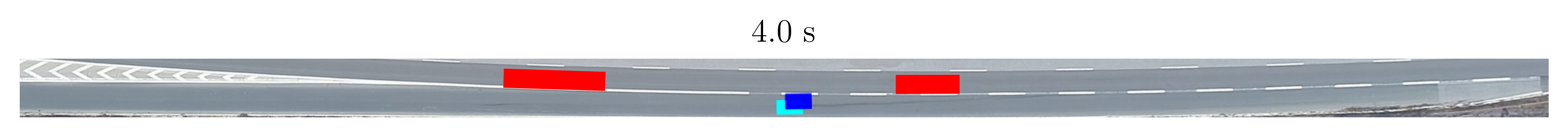}}
    {\includegraphics[width = \textwidth]{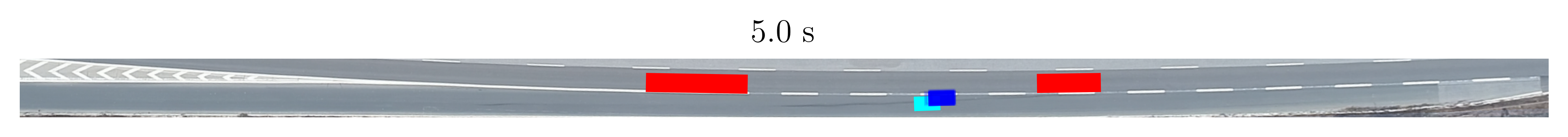}}
    {\includegraphics[width = \textwidth]{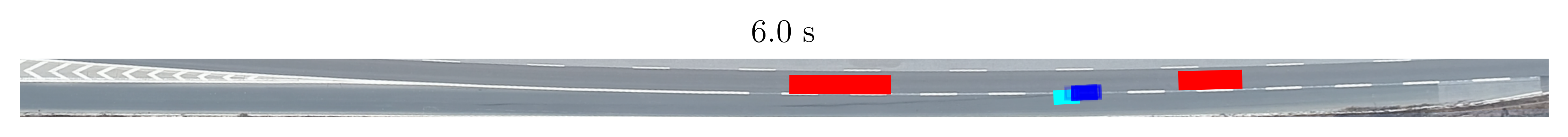}}
    {\includegraphics[width = \textwidth]{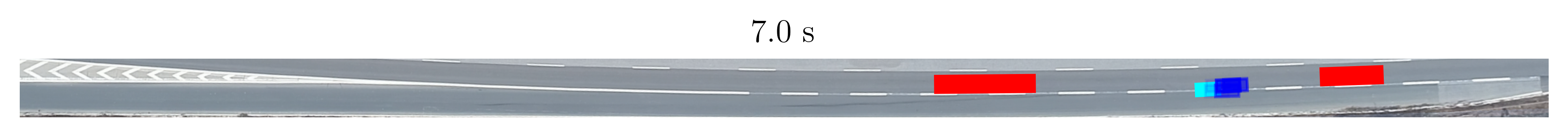}}
     {\includegraphics[width = \textwidth]{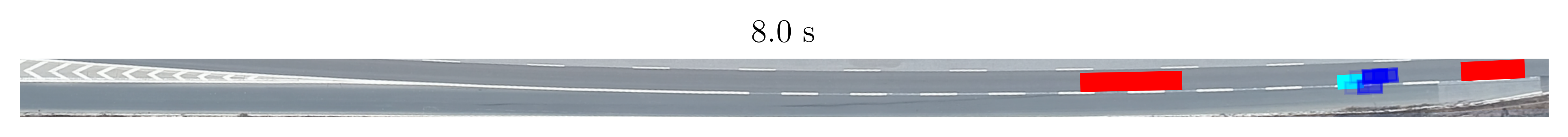}}
      {\includegraphics[width = \textwidth]{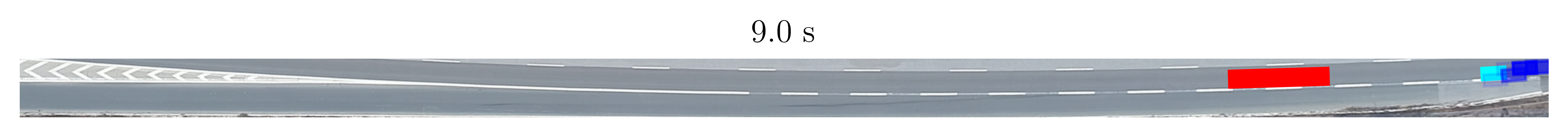}}
  \caption{Time-lapse of the 30 simulations for Scenario 2. Simulated ego vehicles are represented as semi-transparent blue rectangles, non-ego vehicles are shown as red rectangles. The cyan rectangle shows how a real vehicle we replaced with our ego vehicle performed the merge.}
  \label{fig:timelapse2}
 \end{figure*}

\begin{figure*}[!p]
  \centering
   {\includegraphics[width =  \textwidth]{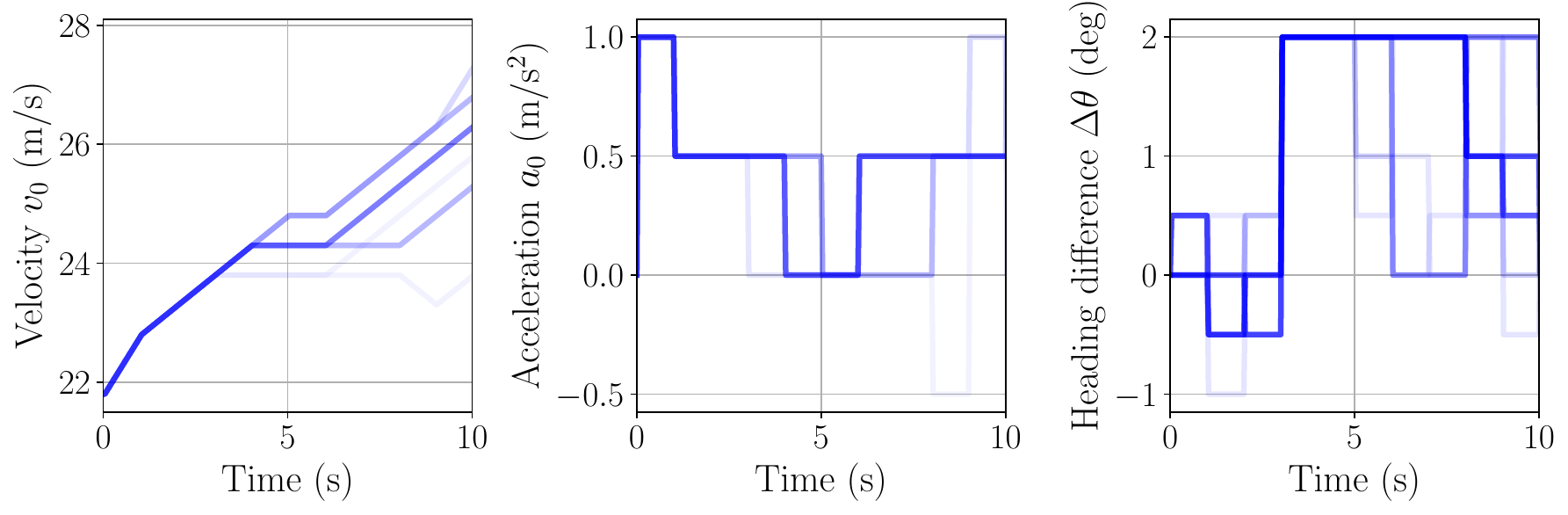}}
  \caption{Velocity and input values of 30 simulations for Scenario 2. The opacity of each line segment represents the total number of simulations sharing the values.}
  \label{fig:values2}
 \end{figure*}

In the second scenario, the ego vehicle starts merging at a higher speed, around 22 m/s. The time-lapse in Figure~\ref{fig:timelapse2} shows that the ego vehicle successfully completed the lane change in all 30 simulations within approximately 8 seconds. This time, the ego vehicle closely mimicked the behavior of the original car. As illustrated in Figure~\ref{fig:values2}, the ego vehicle exhibited gradual acceleration during the merge.

  \begin{figure*}[p]
  \centering
   {\includegraphics[width = \textwidth]{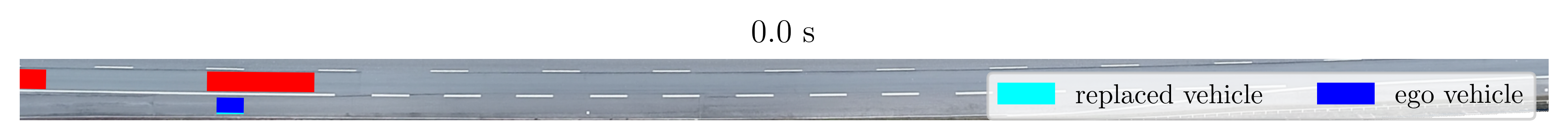}}
   {\includegraphics[width = \textwidth]{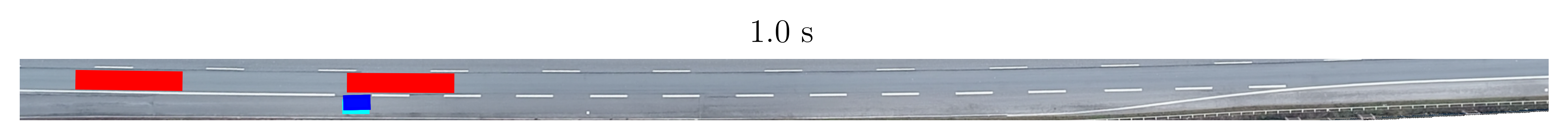}}
   {\includegraphics[width = \textwidth]{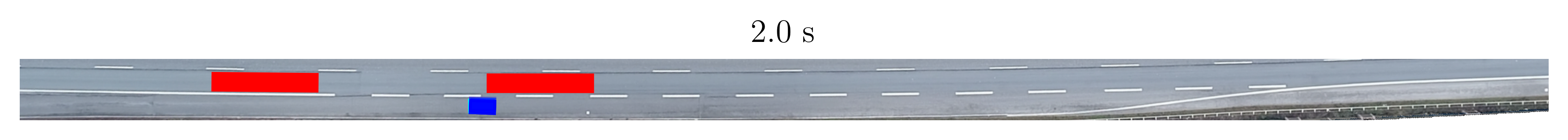}}
   {\includegraphics[width = \textwidth]{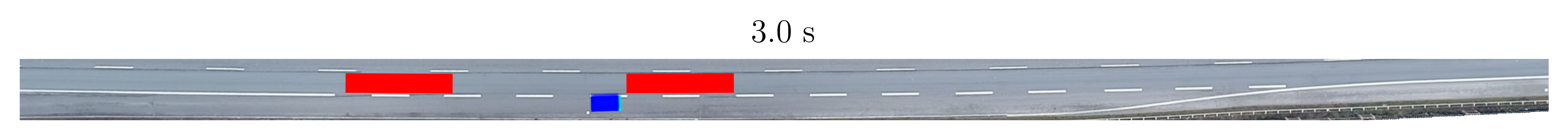}}
   {\includegraphics[width = \textwidth]{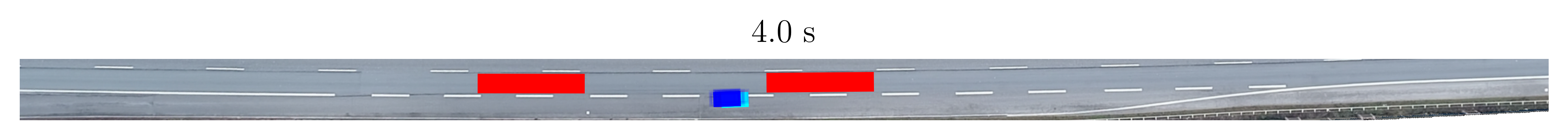}}
       {\includegraphics[width = \textwidth]{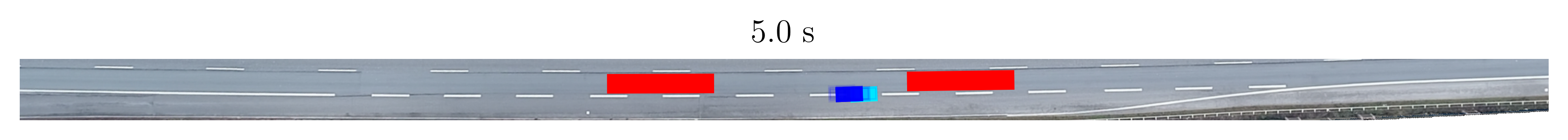}}
       {\includegraphics[width = \textwidth]{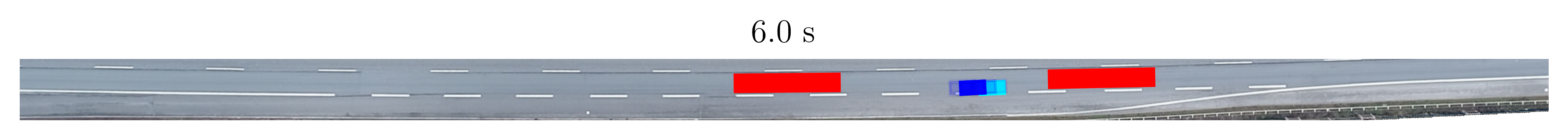}}
       {\includegraphics[width = \textwidth]{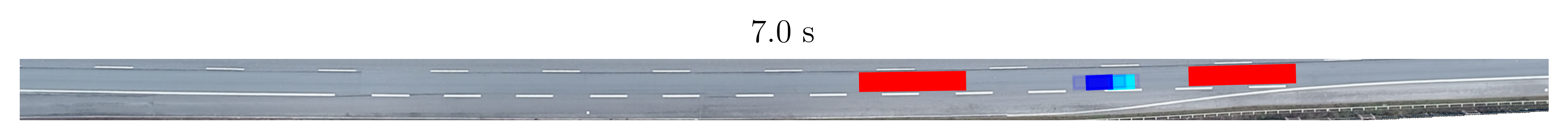}}
       {\includegraphics[width = \textwidth]{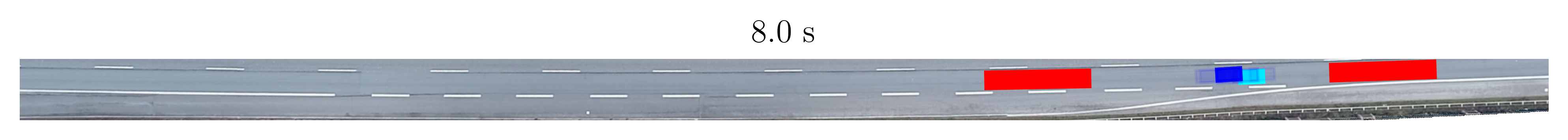}}
        {\includegraphics[width = \textwidth]{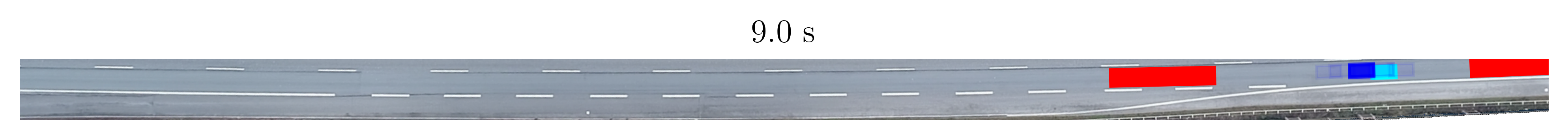}}
  \caption{Time-lapse of the 30 simulations for Scenario 3. Simulated ego vehicles are represented as semi-transparent blue rectangles, non-ego vehicles are shown as red rectangles. The cyan rectangle shows how a real vehicle we replaced with our ego vehicle performed the merge.}
  \label{fig:timelapse3}
 \end{figure*}

  \begin{figure*}[!p]
  \centering
   {\includegraphics[width =  \textwidth]{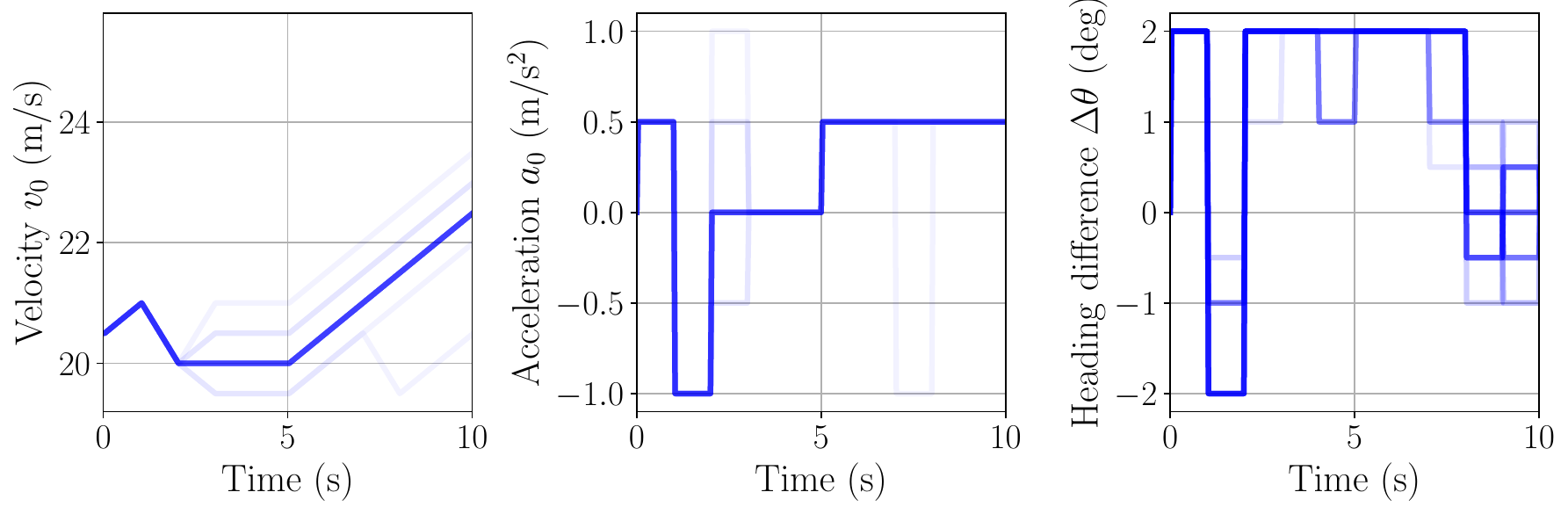}}
  \caption{Velocity and input values of 30 simulations for Scenario 3. The opacity of each line segment represents the total number of simulations sharing the values.}
  \label{fig:values3}
 \end{figure*}
 In the third scenario, the ego vehicle starts at an initial speed of 20.5 m/s. Unlike the previous scenarios, the ego vehicle does not accelerate during the maneuver, as it would not have been feasible to achieve a velocity high enough to merge in front of the adjacent vehicle before the end of the merge lane. The time-lapse in Figure \ref{fig:timelapse3} and the velocity/acceleration graphs in Figure \ref{fig:values3} confirm this behavior. The ego vehicle maintains its velocity, allowing the faster adjacent vehicle to pass, and completes the merge successfully in all simulations within 8 seconds.
 
\section{Conclusion}
\label{sec:conclusion}
This paper presents an extension of our previous work on POMDP-based trajectory planning \cite{vehits24} by adapting the method to the problem of highway on-ramp merging. We demonstrated how this problem can be effectively modeled as a POMDP, integrating vehicle dynamics, measurements, and probabilistic decision-making to handle the uncertainties and interactions present in real-world highway scenarios.

Our simulation results, based on three distinct highway merging scenarios using real-world traffic data from the ExiD dataset, show that the proposed method consistently generates safe and efficient merging maneuvers. The approach successfully handles varying traffic conditions, including different vehicle speeds. 

This work highlights the effectiveness and versatility of POMDP-based planning in addressing complex challenges in automated driving. Looking ahead, further investigation into the real-time capabilities of the method will be crucial for real-world deployment. Additionally, future work will focus on scenarios where assertive driving is critical, such as interactions with other drivers in dense traffic during rush hours. While the POMDP framework is well-suited for these challenges, modeling such interactions and moving beyond offline data evaluation to more interactive testing will be essential to capture these dynamic environments fully.

\begin{credits}
\subsubsection{\ackname} 
This work was co-funded by the European Union under the project ROBOPROX (reg. no. CZ.02.01.01\/00/22\_008\/0004590) and by the Technology Agency of the Czech Republic under the project Certicar CK03000033.
\end{credits}

%
%
%
\bibliographystyle{splncs04}
\bibliography{lib}
%




\end{document}